\pdfoutput=1

\documentclass[11pt]{article}

\usepackage{ACL2023}

\usepackage{times}
\usepackage{latexsym}

\usepackage[T1]{fontenc}

\usepackage[utf8]{inputenc}

\usepackage{microtype}

\usepackage{inconsolata}

\usepackage{xcolor, colortbl} 
\usepackage{graphicx}
\usepackage{xspace}
\usepackage{diagbox}
\usepackage{amsmath}
\usepackage{amsfonts}
\usepackage{subcaption}
\usepackage{natbib}
\usepackage{epsfig}
\usepackage{mathrsfs}
\usepackage{booktabs}  
\usepackage{algpseudocode}
\usepackage{arydshln}
\usepackage{makecell}
\usepackage{multicol}  
\usepackage{multirow}  
\usepackage{CJKutf8}
\usepackage{bm}
\newcommand{\cats}{\textsc{Cats}\xspace}
\newcommand{\catsd}{\textsc{Cats-D}\xspace}
\newcommand{\catss}{\textsc{Cats-S}\xspace}
\newcommand{\coverage}{\textsc{Coverage}\xspace}
\title{\cats: A Pragmatic Chinese Answer-to-Sequence Dataset with Large Scale and High Quality}

\author{Liang Li$^{1, 2}$, Ruiying Geng$^{3}$, Chengyang Fang$^{1, 2}$, Bing Li$^{1}$, Can Ma${}^1$\thanks{$^{*}$Corresponding authors: Can Ma, Yongbin Li}, \\ \textbf{Rongyu Cao}$^{3}$\textbf{,} \textbf{Binhua Li}$^{3}$ \textbf{,} \textbf{Fei Huang}$^{3}$\textbf{,} \textbf{Yongbin Li}$^{3*}$ \\
$^1$Institute of Information Engineering, Chinese Academy of Sciences, Beijing, China \\
$^2$School of Cyber Security, University of Chinese Academy of Sciences, Beijing, China\\
$^3$DAMO Academy, Alibaba Group \\
\texttt{\{liliang, macan\}@iie.ac.cn} \\
\texttt{\{ruiying.gry, shuide.lyb\}@alibaba-inc.com}
}

\begin{document}
\maketitle
\begin{abstract}
There are three problems existing in the popular data-to-text datasets. First, the large-scale datasets either contain noise or lack real application scenarios. Second, the datasets close to real applications are relatively small in size. 
Last, current datasets bias in the English language while leaving other languages underexplored.
To alleviate these limitations, in this paper, we present \cats, a pragmatic Chinese answer-to-sequence dataset with large scale and high quality. The dataset aims to generate textual descriptions for the answer in the practical TableQA system.
Further, to bridge the structural gap between the input SQL and table and establish better semantic alignments, we propose a Unified Graph Transformation approach to establish a joint encoding space for the two hybrid knowledge resources and convert this task to a graph-to-text problem. The experiment results demonstrate the effectiveness of our proposed method. Further analysis on \cats \footnote{\cats is available at \url{https://github.com/AlibabaResearch/DAMO-ConvAI/tree/main/cats}} attests to both the high quality and challenges of the dataset.
\end{abstract}

\section{Introduction}
Data-to-text (D2T) generation \citep{DBLP:conf/acl/Kukich83, DBLP:journals/nle/ReiterD97} aims to generate a natural language description conditioned on structured or semi-structured data, such as graphs~\citep{song2018graph,wang2020better} or tables~\citep{lebret2016neural,wiseman2017challenges}. It helps people get the key points of the input data and makes the stored information accessible to a broader range of end-users. A large number of datasets have been proposed as the testbed for neural D2T models and are driving the domain.

However, as shown in Table~\ref{tab:data2text_dataset_comparision}, we note three problems existing in the popular datasets. 
First, the large-scale datasets either contain noises (e.g., WEATHERGOV~\citep{liang2009learning}) or lack practical application scenarios, e.g., ToTTo~\citep{parikh2020totto}. The shortcoming leads to a separation between research and application. Second, the datasets close to practical scenarios are relatively small in size. For example, ROTOWIRE~\citep{wiseman2017challenges} only contains 4.9K training examples, and CoSQL~\citep{yu2019cosql} is consist of 7.8K training pairs. The small training size can easily lead to overfitting and is not conducive to training a reliable neural network model. 
Lastly, most of the existing datasets are built for English, which leads to advanced work on D2T generation primarily focusing on English and leaving other languages underexplored. 
These limitations hinder the progress of D2T generation. We therefore need to investigate possible remedies.

\begin{figure}[t]
\centering
\includegraphics[width=1.0\columnwidth]{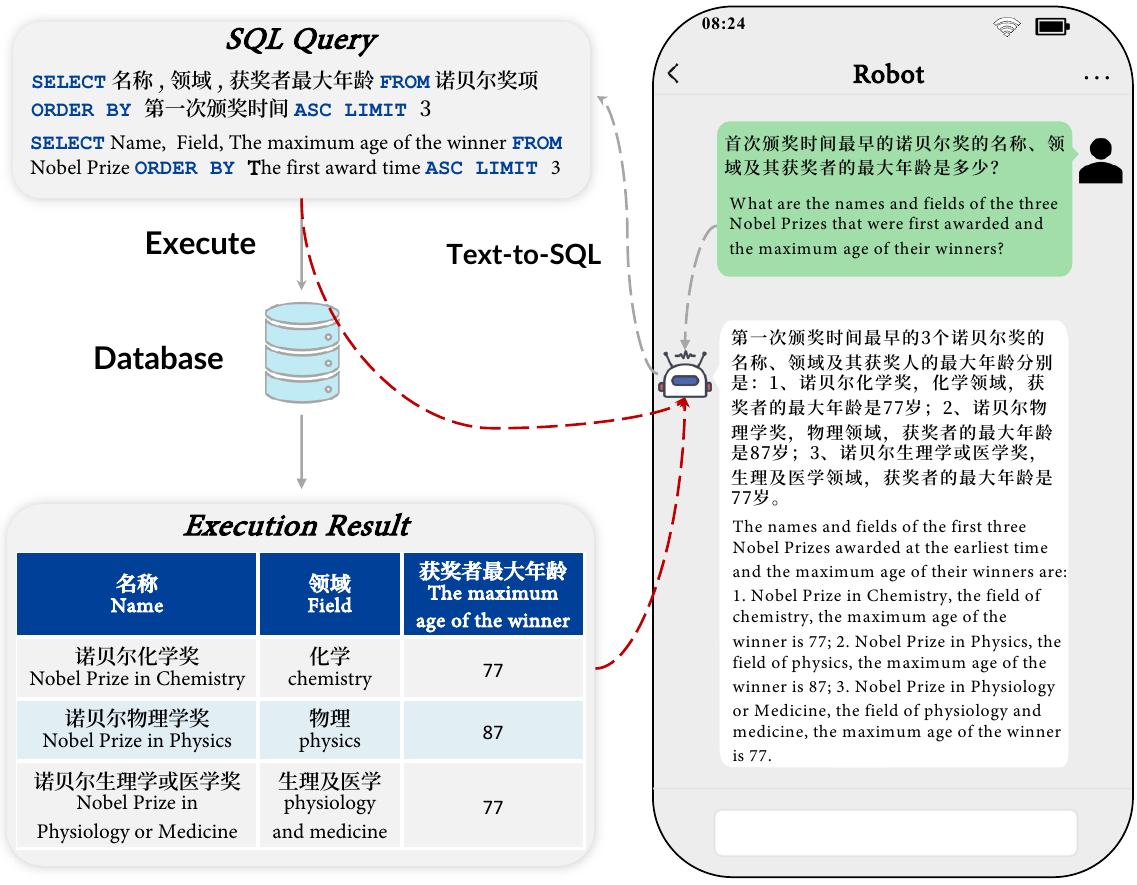}
\caption{An example for a practical TableQA system. The red dotted lines denote the input data  for answer-to-sequence.}
\label{fig1:introduction}
\end{figure}

The crucial step to improving the above limitations is digging out a data-to-text task with a practical scenario.
Recently, CoSQL \citep{yu2019cosql} has proposed a practical controlled D2T task: answer-to-sequence. As shown in Figure \ref{fig1:introduction}, the task takes a SQL query generated by a semantic parsing module, i.e., text-to-SQL~\citep{DBLP:journals/corr/abs-1207-1420}, and its corresponding execution result (in the form of a table) as the model input and aims to produce a natural language description as the response to users in a real-world TableQA system. The SQL gives  explicit signals for models on what to generate.
The generated description could provide a concise and easy-to-understand summary of the result table and help users verify whether the queried result is consistent with the original question~\citep{fang2022faithful}.
Moreover, the task also contributes to a more user-friendly human-computer interaction. 
Nevertheless, CoSQL contains only 7.8K answer-to-sequence examples for training. Additionally, it is a dataset with SQL-grounded dialogue state tracking as the core, and the generation annotations are very rough. 
The scale and quality of CoSQL limit further exploring the answer-to-sequence task.

In this paper, to bridge the gap between research and application of data-to-text datasets and enrich their language diversity, we comply with the CoSQL setting and present \cats, a large-scale and high-quality \underline{\bf C}hinese \underline{\bf a}nswer-\underline{\bf t}o-\underline{\bf s}equence dataset.
We manually annotate all collected SQL-table pairs to obtain their descriptions. We make two efforts to improve the quality and scale of the collected SQL-Table pairs and guarantee they are close to practical scenarios.
First, we annotate the SQL-table pairs from DuSQL~\citep{wang-etal-2020-dusql}, a large-scale Chinese Text-to-SQL dataset with a SQL query distribution close to real applications.
Data collected in this way are named \catsd.
Second, we adopt an automatic data construction pipeline to collect a large number of SQL-table pairs for annotation. The basic idea is automatically crawling a mount of tables from the Internet to build multi-table databases and then automatically generating SQL queries based on the SQL grammar and constrained by the given database. Data collected with this method are referred to as \catss.
Compared to \catsd, \catss expands the data scale while reducing the share of easy SQLs to make the dataset more challenging.
In total, \cats is made up of both \catsd and \catss, and contains 43,369 answer-to-sequence examples, which is an order of magnitude larger than CoSQL.

The input SQL and table in answer-to-sequence are heterogeneous, and there is a structural gap between them. To bridge the gap and establish better semantic alignments, we propose a Unified Graph Transformation approach (\textproc{Ugt}), which first converts the two sources to two undirected graphs, then builds the connection between the nodes in different graphs to obtain a unified graph. In this way, we convert this task to a graph-to-text problem~\citep{DBLP:conf/inlg/GardentSNP17}. Previous graph-to-text work \citep{ribeiro-etal-2021-structadapt} transforms the input graph into a new token graph to apply pretrained language models, such as T5~\citep{2020t5}. We consider that this transformation breaks the original input graph structure and may bring in extra noises into graph encoding. Hence, we further introduce a Node Segment Embedding (\textproc{Nse}) to preserve original structure information.

Our contributions are three-fold as follows: 
\begin{itemize}
    \setlength{\itemsep}{1pt}
    \setlength{\parsep}{1pt}
    \setlength{\parskip}{1pt}
\item We present a large-scale and high-quality Chinese answer-to-sequence dataset (\cats), which narrows the gap between research and application of data-to-text generation datasets and enriches the language diversity.
\item We propose \textproc{Ugt} and \textproc{Nse} to better model the input of two heterogeneous structured input data sources.
\item Experiments and analysis on \cats attest to both the high quality and challenges of the dataset.
The results also demonstrate the effectiveness of our proposed method.
\end{itemize}

\begin{table*}[h!]
    \centering
    \resizebox{\textwidth}{!}{
    \begin{tabular}{lrllll}
    \toprule
    \textbf{Dataset} & \textbf{Train Size} & \textbf{Domain} & \textbf{Target} & \textbf{Application} & \textbf{Language} \\
    \midrule
    WEATHERGOV~\citep{liang2009learning} & 25K & Weather & Crawled & Weather Report & English \\
    WikiBio~\citep{lebret2016neural} & 583K & Wikipedia & Crawled & - & English \\
    WebNLG~\citep{gardent2017creating} & 25.3K & DBPedia & Annotated & - & English \\
    LogicNLG~\citep{DBLP:conf/acl/ChenCSCW20} & 28.5K & Wikipedia & Annotated & - & English \\
    ToTTo~\citep{parikh2020totto} & 120K & Wikipedia & Annotated & - & English \\
    \midrule
    Rotowire~\citep{wiseman2017challenges} & 4.9K & NBA & Annotated (Noisy) & NBA & English \\
    AdverGeneration~\citep{DBLP:conf/emnlp/ShaoHWXZ19} & 115K & Chinese E-commerce & Crawled & Advertising Text Generation & Chinese \\
    CoSQL~\citep{yu2019cosql} & 7.8K & Cross-Domain & Annotated & TableQA & English \\
    Map2seq~\citep{DBLP:conf/acl/SchumannR20} & 7.6K & OpenStreetMap & Annotated & Navigation & English \\
    \midrule
     \rowcolor[RGB]{237,237,237} \cats & 34.7K & Cross-Domain & Annotated & TableQA & Chinese \\
     \rowcolor[RGB]{237,237,237} \catsd & 6.7K & Cross-Domain & Annotated & TableQA & Chinese \\
     \rowcolor[RGB]{237,237,237} \catss & 26.4K & Cross-Domain & Annotated & TableQA& Chinese  \\
    \bottomrule
    \end{tabular}
    }
    \caption{Comparison of popular data-to-text datasets in different aspects. \textbf{Application} represents the practical application scenario associated with the dataset.}
    \label{tab:data2text_dataset_comparision}
\end{table*}

\section{Related Works}
\subsection{Answer-to-Sequence Generation}
In a real-world setting, a TableQA system comprises a table semantic parsing (text-to-SQL) component and an answer-to-sequence component. The semantic parsing component converts a natural language question into a SQL query \citep{guo2019towards, wang2020rat, hui2021dynamic} and the answer-to-sequence component aims generating a natural language description of the SQL and the execution result. 
CoSQL~\cite{yu2019cosql} first proposes the answer-to-response task and refers to it as response generation. Intuitively, response generation should encompass both answer acquisition and answer description, which could easily be confused with the role of the whole Table QA system. Therefore, to make the task more clearly related to its definition and function, we rename it as answer-to-sequence generation.
In this paper, the proposed \cats follows the same task setting in CoSQL. Specifically, the task's input consists of a SQL query and its corresponding execution result (in the form of a table), and the output is a natural language description.
Especially, using SQL query as input rather than natual language question is more practical in multi-turn TableQA scenarios because the SQL query can easily represent the context state \citep{yu2019cosql}.

\subsection{Structure Modeling in Data-to-Text}
Recently, some works in D2T generation have shown that the structure modeling for the input data can dramatically improve the model performance.
For table data, \citet{liu2019hierarchical,li2021improving} propose to utilize a hierarchal encoder to model the table's representation from the row and column levels.
For graph structure modeling, early works \citep{song2018graph,damonte2019structural} introduce Graph Neural Networks as the structure encoder, which only considers the relations between neighbor nodes. Unlike the local encoding strategies, \citet{zhu2019modeling,cai2020graph} propose the Graph Transformer that uses explicit relation encoding and allows direct communication between two distant nodes. 
Newly, some works enable the pretrained language models the structure modeling capabilities and achieve SOTA results on many D2T tasks. Especially, \citet{ribeiro-etal-2021-structadapt} attempt to insert structural adapters into T5'encoder to model the graph structure. \citet{DBLP:conf/naacl/WangXSC22} modify the T5's attention masking matrix to encode table with a structure-aware self-attention mechanism. 
In this paper, we propose to utilize \textproc{Ugt} to convert the input SQL and table to a graph and utilize a graph-to-model to model it. Our model refers to \citet{ribeiro2020modeling,ribeiro-etal-2021-structadapt}' works and further improves them by introducing \textproc{Nse} to better preserve the graph structure.

\section{Dataset Construction}
\label{data_construction}
Considering the limitations of existing D2T datasets, we present \cats, a massive and pragmatic Chinese answer-to-sequence dataset. 
\cats is constructed by two phases: SQL-table pairs collection and manual data annotation. 
To balance the data quality and scale and bring it closer to the practical scenario, we collect the SQL-table pairs in two ways. 
First, we derive SQL-table pairs from DuSQL~\citep{wang-etal-2020-dusql}, a text-to-SQL dataset that generates the SQL queries by referring to the SQL query distribution in real-life applications. The dataset obtained by annotating these pairs is referred to as \catsd.
Besides, we implement an automatic data construction pipeline to collect massive high-quality SQL-table pairs. 
Data collected with this method are referred to as \catss, which increases the proportion of complicated SQL queries to make the dataset more challenging.
Ultimately, both \catsd and \catss make up \cats.
We first describe how to obtain SQL-table pairs for subsequent annotation and then introduce the annotation details.

\paragraph{Database Building}
To mimic the practical TableQA system, we first follow~\citet{wang-etal-2020-dusql} to build a multi-table database $D^d$ by collecting all databases in DuSQL.
In addition, we also build another multi-table database $D^s$ for expanding the size and domain of our dataset through a table collection pipeline.
Specifically, 100,000 high-frequency words are first summarized from the CLUE~\citep{xu2020clue} corpus.
Then, we query these words in Google and download all the queried spreadsheet files. Subsequently, the available tables in these spreadsheets are extracted by a table parser that can identify the potential table in a worksheet.
To protect personal privacy, we use predefined unique words to replace sensitive information in these tables, such as passwords, ID numbers, credit card numbers, etc. Finally, these tables are used to construct the database $D^s$. Please refer to Appendix~\ref{table_clean} for more details.

\paragraph{SQL and Table Collection}

We execute all the SQL queries in DuSQL in the database $D^d$ to get their corresponding tables. This is consistent with how a practical Table QA system answers user questions after parsing it to SQL. 
Then we discard SQL-table pairs containing SQLs that execute with empty results to obtain a SQL-table pair set CATS-D$^{un} = \{s^d_i, t^d_i\}_{i=1}^n$. 
DuSQL does not release the code for generating synthetic queries. Therefore, to increase the annotation examples, we reimplement a SQL generator similar to the one in DuSQL.
Notably, the generated SQL contains both single-table and multi-table queries.
Please refer to Appendix~\ref{appendixsec:sql_generator} for more detailed information on the SQL generator.
The sampled SQLs which cannot execute in database $D^s$ or execute with empty results are deserted. In this way, we obtain another SQL-table pair set CATS-S$^{un} = \{s^s_i, t^s_i\}_{i=1}^m$.

\paragraph{Data Annotation Process}
We employ 20 well-educated crowd workers to annotate the SQL-table pairs in CATS-D$^{un}$ and CATS-S$^{un}$. In particular, the annotators are asked to write a description $y$ given a SQL $s$ and table $t$ pair. They must follow the requirements: (1) avoiding template-like language and trying to write a natural, fluent, and grammatically correct description; (2) the description must summarize all the content in the table; (3) the description must be logically consistent with the input SQL; (4) filtering the incomprehensible examples that are semantically unclear.
Furthermore, to guarantee data quality, another 4 workers are asked to review the annotated data. Data with poor annotation quality will be required to be relabeled. 
Finally, the annotated CATS-D$^{un}$ is named as \catsd. 
To guarantee data consistency, we sample a subset from the annotated CATS-S$^{un}$ following a similar complexity distribution with \catsd.
We name the sampled dataset \catss. However, we find that easy SQL queries account for a large-scale proportion (\textbf{47.87\%}) in \catsd. Therefore, we reduce the proportion of easy SQLs (\textbf{14.50\%}) in \catss to make it more challenging.

\subsection{Dataset Analysis}
The final \cats contains 43,369 examples, including 8,350 examples in \catsd and 33,019 examples in \catss. Each annotated example contains a triple of SQL $s$, table $t$, and descriptive sentences $y$. We split the training/development/test sets by 34,697/4,336/4,336 randomly. To understand the characteristics of the data collected in \catsd and \catss, we also split them accordingly. The training, development, and test sets of \catsd and \catss contain 6,680/835/835 and 28,017/3,501/3,501 examples, respectively.

\begin{table}[t!]
    \centering
\resizebox{\columnwidth}{!}{
\begin{tabular}{lrrrr}
\toprule
\bf \textsc{Column Number} & 1& 2 & 3 & $>=$4 \\
 \midrule
 CoSQL & 6,329 & 1057 & 459 & 0 \\
 \cats & 8,966 & 20,862 & 3242 & 1627 \\
 \catsd & 2,883 & 2,977 & 820 & 0   \\
\catss & 6,157 & 17,813 & 2,394 & 1,653  \\
\midrule
 \bf \textsc{Row Number} & 1 & 2 & 3 & $>=$4 \\
\midrule
CoSQL & 4740 & 610 & 2,495 & 0  \\
\cats & 14,909 & 6,158 & 3,671 & 9,959  \\
\catsd & 2,123 & 656 & 1,129 & 2,772  \\
\catss & 12,754 & 5,538 & 2,510 & 7,215  \\
\midrule
 \bf \textsc{SQL Hardness} & Easy & Medium & Hard & Extra Hard \\
\midrule
CoSQL & 2,788 & 1,826 & 1,717 & 1,514 \\
\cats & 7,223 & 13,000 & 12,016 & 2,458 \\
\catsd & 3,198 & 1709 & 1,264 & 509  \\
\catss & 4,063 & 11,214 & 10,787 & 1,953  \\
 \midrule

 \bf \textsc{Target Length} & $<20$ & $< 40$ & $<60$ & $>=60$ \\
\midrule
CoSQL & 7,005 & 825 & 15 & 0\\
\cats & 10,319 & 12,862 & 5,864 & 5,652\\
\catsd & 1,893 & 2,026 & 1,912 & 849 \\
\catss & 8,401 & 10,873 & 3,962 & 4,781  \\
\bottomrule
\end{tabular}
}
    \caption{Complexity distribution comparison between \cats and CoSQL. }
    \label{tab:data_distribution_comp}
\end{table}
\begin{figure*}[!ht]
\centering
\includegraphics[width=1.0\textwidth]{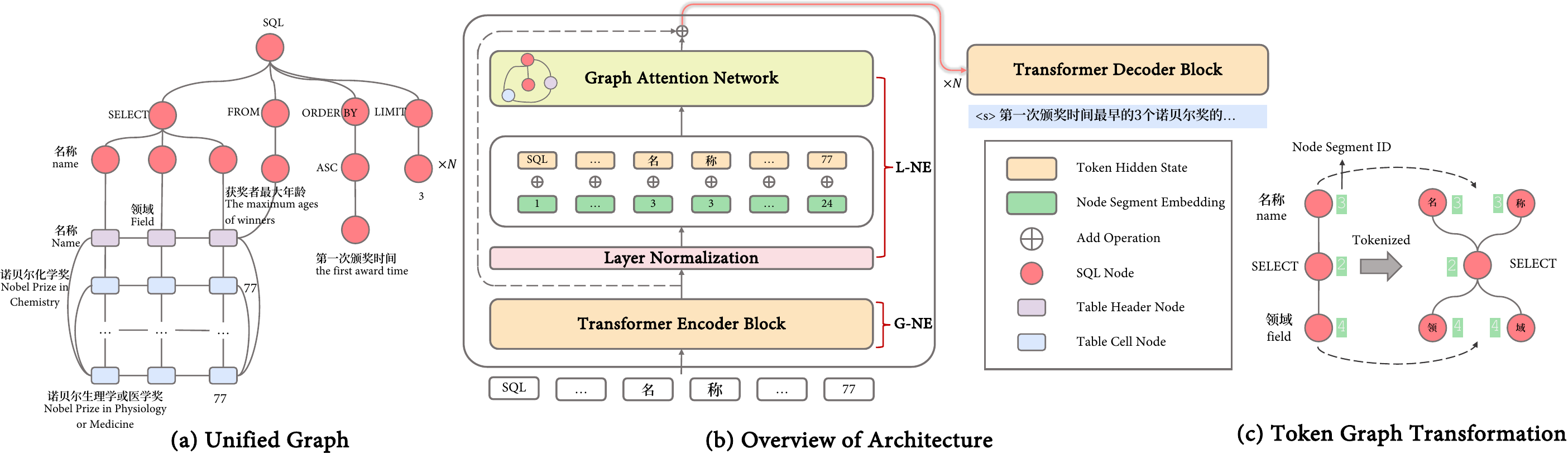} 
\caption{Illustration of the proposed method. (a) is an example of a unified graph transformed by Unified Graph Transformation. (b) is an overview of our model. \textbf{L-NE} and \textbf{G-NE} denote Local Node Encoder and Global Node Encoder, respectively. (c) is an example of token graph transformation.}
\label{fig2:method}
\end{figure*}
\paragraph{Data Complexity} 
To better understand our dataset, we compare its complexity with CoSQL in four dimensions, including the input tables' row and column number, SQL hardness, and the target length. 
Following~\citet{guo2021chase}, we adopt SQL hardness to measure the complexity of SQL queries from the following four-level: easy, medium, hard, and extra hard, according to the number of components, selections, and conditions in a SQL query \citep{yu2018spider}. Considering CoSQL only release the training and delvelopment sets, we only show the training set comparision. The results are summarized in Table~\ref{tab:data_distribution_comp}. 
First, we find that the tables in CoSQL are small, such as 60\% of the tables with only one row and more than 80\% with only one column.
Second, we notice that most of the descriptions in CoSQL are less than 20 in length. The first reason is that most of the input tables are small. By manually checking the data in CoSQL, we find the second reason is that CoSQ describes the table with more than two rows through a generic template, such as ``\verb|Here are the ...|''. 
Last, we observe that easy SQL queries in CoSQL account for \textbf{35.54\%}, far more than \textbf{20.84\%} in \cats.
These features make CoSQL only suitable for simple scenarios and less challenging.
By contrast, \cats has a broader distribution than CoSQL, which is more in line with real TableQA applications.

\section{Structure-Aware Approach}
Given an input SQL $s$ and a table $t$, the model aims to generate a response $\tilde{y}$.
To bridge the gap between the two sources of information, we first propose a \textbf{Unified Graph Transformation} approach (UGT), which explicitly connects the input SQL and table in a unified structure. In this way, we can obtain a joint graph representation of the two sources and convert the answer-to-sequence task to a graph-to-text problem. And then, we utilize a varietal transformer architecture \citep{ribeiro2020modeling} that employs the original transformer encoder as the Global Node Encoder (G-NE) and introduces a GNN based layer into each transformer encoder layer as the Local Node Encoder (L-NE). G-NE allows explicit communication between two distant nodes, taking advantage of a large node context range. And L-NE has an advantage in modeling the graph topology. As shown in Figure \ref{fig2:method} (b), this architecture cascaded performs global and local node aggregation, which gathers the benefits from both strategies. In the rest of this section, we will describe the proposed Unified Graph Transformation and the Local Node Encoder in detail.

\subsection{Unified Graph Transformation}
\label{subsec: unified}
Given a SQL $s$ and its execution result (in the form of a table) $t$ as input (shown in Figure \ref{fig1:introduction}), the Unified Graph Transformation takes two steps to transform the input two sources of data into a unified graph (shown in Figure \ref{fig2:method} (a)). First, it converts the SQL and table into two undirected graphs: SQL graph $\mathcal G_s$ and table graph $\mathcal G_t$. In particular, for a SQL, we follow the previous method \citep{xu2018graph2seq} and convert it to a tree. 
For a table, we treat each column name and table cell as a node and divide the nodes in the table into two categories: table header node and table cell node. And then, we connect each header node with the cell node in the same column. We also build the connections between the cell nodes in the same row. Second, we add connections between the nodes that indicate the same column in $\mathcal G_s$ and $\mathcal G_t$ to build the unified graph.
we also add a self-loop connection for each node. The transformed unified graph is formulated as $\mathcal G_h = ( \mathcal V_h, \mathcal E_h) $, where $\mathcal V$ represents the nodes set and $\mathcal E_h = \{ (n, v) | n,v \in \mathcal V \}$. Figure \ref{fig2:method} (a) shows an example of the transformed unified graph.

We expect that developing generation model should benefit from the recent advance on pretrained language models (PLMs). Following previous work \citep{ribeiro-etal-2021-structadapt}, we represent each $\mathcal G_h$ using subword tokens, and convert it into a new token graph $\mathcal G = (\mathcal V, \mathcal E)$. Specifically, each token of a node in $V_h$ becomes a node $\tilde v$ in $\mathcal N$.  For each edge $(n, v) \in \mathcal E_h$, we connect each token between $n$ and $v$ to obtain the new edges set $\mathcal E$ (as shown in Figure~\ref{fig2:method} (c)). However, we notice that the new token graph $\mathcal G$ breaks the structure of the original graph $\mathcal G_h$ and may make the encoder pay too much attention to the feature of nodes at the token level instead of the original node level. This may bring extra noise into graph encoding. To preserve the original structural information, we introduce the \textbf{Node Segment Embedding} (NSE), which assigns the same symbol to the nodes in the token graph $\mathcal G$ which belong to the same node in the original unified graph $\mathcal G_h$. Figure~\ref{fig2:method} (c) gives an example.

\subsection{Local Node Encoder}
Given $\{ h_v | v \in \mathcal V \}$ as the outputs of the Global Node Encoder at the $L$-th encoder layer, we next describe how the Local Node Encoder (L-NE) works. As shown in Figure \ref{fig2:method} (b), L-NE consists of two main modules: a Node Embedding Layer and a Graph Attention Network (GAT) \citep{DBLP:conf/iclr/VelickovicCCRLB18} Layer. The former enriches the features of the nodes, and the latter explicitly models the graph structure. Formally, given $h_v$, we obtain the feature-enhanced node representation by:
\begin{align}
  h^e_v &= LayerNorm(h_v) + e^s_v,  
\end{align}
where $LayerNorm$ represents layer normalization \citep{ba2016layer}. $e^s_v$ denote the node segment embedding for node $v$. 
After the Node Embedding Layer, we utilize a GAT layer to model the graph structure. Formally, it aggregates the representations of node $v$ in a multi-head self-attention layer \citep{vaswani2017attention} as follows:

\begin{equation}
\resizebox{.5\hsize}{!}{$
\begin{aligned}
s^h_{v,n} &= \frac{h^e_v W^h_Q (h^e_n W^h_K)^\top }{ \sqrt{d/H} }, \\
\alpha^h_{v,n} &= \frac{ e^{ s^h_{v,n} } }{\sum_{ \tilde{n} \in \mathcal N(v)} e^{ s^h_{v, \tilde{n}} } }, \\
z^h &=  \sum_{ n \in \mathcal N(v)} \alpha^h_{v,n} (h^e_n W^h_V), \\
h^r &= Concat(z^1, ..., z^H),
\end{aligned}
$}
\end{equation}
where $1 \le h \le H$, and $W^h_Q$, $W^h_K$, $W^h_V \in \mathbb{R}^{d \times (d/H)} $. $\mathcal N(v)$ denotes the
immediate neighborhood of node $v$ in graph $\mathcal G$. 

The transformer parameters are initialized with the pretrained T5 \citep{2020t5}, and the others are randomly initialized. 
Given each gold instance $(s, t, y)$, we fine-tune the model to optimize the following cross-entropy objective:
\begin{align}
\label{equation_language_modeling}
    \mathcal L &= -\sum^{|y|}_{i=1} p_{\theta} ( y_i|y_{1:i-1} ; s, t).
\end{align}

\section{Experiment}
\subsection{Experiment Settings}
\begin{table}[t]
\centering
\small
    \centering
    \begin{tabular}{ll}
        \toprule
        \textbf{SQL Components} &  \textbf{Descriptions}\\
        \midrule
        \texttt{Min} & \begin{CJK}{UTF8}{gbsn} 最小的 (minimum) \end{CJK} \\
        \texttt{Max} & \begin{CJK}{UTF8}{gbsn} 最大的 (maximum) \end{CJK} \\
        \texttt{Count} & \begin{CJK}{UTF8}{gbsn} 数量 (the number of) \end{CJK} \\
        \texttt{Sum} & \begin{CJK}{UTF8}{gbsn} 总共 (total) \end{CJK} \\
        \texttt{Average} & \begin{CJK}{UTF8}{gbsn} 平均 (average) \end{CJK} \\
        \midrule
        \texttt{=} & \begin{CJK}{UTF8}{gbsn} 等于 (is) \end{CJK} \\
        \texttt{!=} & \begin{CJK}{UTF8}{gbsn} 不等于 (is not) \end{CJK} \\
        \texttt{>} & \begin{CJK}{UTF8}{gbsn} 大于 (more than) \end{CJK} \\
        \texttt{>=} & \begin{CJK}{UTF8}{gbsn} 大于等于 (no less than) \end{CJK} \\
        \texttt{<} & \begin{CJK}{UTF8}{gbsn} 小于 (less than) \end{CJK} \\
        \texttt{<=} & \begin{CJK}{UTF8}{gbsn} 不小于 (no more than) \end{CJK} \\
        \midrule
        \texttt{And} &  \begin{CJK}{UTF8}{gbsn}并且 (and)\end{CJK} \\
        \texttt{Or} & \begin{CJK}{UTF8}{gbsn}或者 (or)\end{CJK} \\
        \midrule
        \texttt{Asc} & \begin{CJK}{UTF8}{gbsn}从低到高 (in the ascending)\end{CJK} \\

        \texttt{Desc} & \begin{CJK}{UTF8}{gbsn}从高到低 (in the descending)\end{CJK} \\
        \bottomrule 
    \end{tabular}
    \caption{Natual language descriptions for different SQL components.}
    \label{sql_description}
\end{table}
\begin{table*}[!ht]
    \centering
    \resizebox{\textwidth}{!}{
    \begin{tabular}{llllllllll}
    \toprule
        \multirow{2}{*}{\textsc{\textbf{Models}}} & \multicolumn{3}{c}{\textbf{\cats}} & \multicolumn{3}{c}{\textbf{\catsd}} & \multicolumn{3}{c}{\textbf{\catss}} \\
        \cmidrule(lr){2-4} \cmidrule(lr){5-7} \cmidrule(lr){8-10}
         & \textbf{BLEU}  & \textbf{ROUGE-L} & \textbf{\coverage} & \textbf{BLEU}  & \textbf{ROUGE-L} & \textbf{\coverage}& \textbf{BLEU} & \textbf{ROUGE-L} & \textbf{\coverage} \\
    \midrule
    \multicolumn{10}{c}{\em Development} \\
    \midrule
    \textproc{Gold} & - & - & 75.56 & - & - & 69.59 & - & - & 77.30 \\
    \midrule
    \textproc{TemP} & 40.04  & 57.20 & 81.48 & 18.05 & 47.37 & 77.93 & 42.71 & 59.82 & 83.24 \\
    \midrule
    \textproc{Pointer-Gen} 
    & 
    51.26$_{\pm 0.20}$ & 73.70$_{\pm 0.14}$ & 68.73$_{\pm 0.13}$ 
    &
    48.33$_{\pm 0.91}$ & 67.95$_{\pm 0.96}$ & 56.96$_{\pm 0.90}$ 
    &
    49.77$_{\pm 0.16}$ & 73.79$_{\pm 0.26}$& 69.26$_{\pm 0.24}$\\
    \textproc{T5} & 
    53.60$_{\pm 0.13}$ & 74.42$_{\pm 0.06}$ & 72.87$_{\pm 0.04}$ &
    52.47$_{\pm 0.28}$ & 68.5$_{\pm 0.32}$ & 68.20$_{\pm 0.25}$ & 
    51.43$_{\pm 0.10}$ & 73.77$_{\pm 0.04}$ & 73.08$_{\pm 0.03}$ \\
    
    \textproc{T5 + Fnn} & 
    54.14$_{\pm 0.21}$ & 74.80$_{\pm 0.16}$ & 72.85$_{\pm 0.18}$ & 
    52.10$_{\pm 0.17}$ & 68.28$_{\pm 0.17}$ & 68.02$_{\pm 0.31}$ &
    51.67$_{\pm 0.22}$ & 73.75$_{\pm 0.17}$ & 73.08$_{\pm 0.17}$\\
    
    \textproc{T5-Graph} & 
    52.21$_{\pm 0.17}$ & 73.68$_{\pm 0.04}$ & 72.03$_{\pm 0.10}$ & 
    49.89$_{\pm 0.40}$ & 66.72$_{\pm 0.10}$ & 66.65$_{\pm 0.26}$ & 
    50.12$_{\pm 0.18}$ & 73.11$_{\pm 0.13}$ & 72.05$_{\pm 0.04}$\\
    
    \textproc{T5-Graph + Fnn} & 
    52.30$_{\pm 0.17}$ & 73.71$_{\pm 0.20}$ & 71.87$_{\pm 0.05}$ & 
    48.81$_{\pm 0.27}$ & 66.35$_{\pm 0.13}$ & 66.10$_{\pm 0.30}$ & 
    50.42$_{\pm 0.09}$ & 73.22$_{\pm 0.12}$ & 72.07$_{\pm 0.05}$\\
    
    \midrule
    \textproc{Ugt} & 
    54.75$_{\pm 0.15}$ & 75.72$_{\pm 0.06}$ & 72.68$_{\pm 0.16}$ & 
    54.23$_{\pm 0.49}$ & 69.82$_{\pm 0.35}$ & 68.07$_{\pm 0.63}$ & 
    52.54$_{\pm 0.16}$ & 74.84$_{\pm 0.12}$ & 72.99$_{\pm 0.07}$ \\
    
    \rowcolor[RGB]{237,237,237} \textproc{Ugt + Nse} & 
    \bf 56.34$_{\pm 0.13}$ & \bf 76.72$_{\pm 0.09}$ & \bf 73.41$_{\pm 0.05}$ & 
    \bf 58.79$_{\pm 0.51}$ & \bf 73.16$_{\pm 0.31}$ & \bf 68.94$_{\pm 0.31}$ & 
    \bf 53.54$_{\pm 0.15}$ & \bf 75.36$_{\pm 0.19}$ & \bf 73.67$_{\pm 0.10}$ \\

    \midrule
    \multicolumn{10}{c}{\em Test} \\
    \midrule
    \textproc{Gold} & - & - & 76.35 
    &
    - & - & 68.67 
    &
    - & - & 76.98 \\
    \midrule
    \textproc{TemP} 
    &
    41.39 & 57.82 & 82.40 
    & 
    17.76 & 46.21 & 77.83 
    &
    42.69 & 60.16 & 82.96 \\
    \midrule
    \textproc{Pointer-Gen}  
    &
    50.77$_{\pm 0.56}$ & 73.25$_{\pm 0.14}$ & 68.47$_{\pm 0.31}$ 
    & 
    47.34$_{\pm 0.81}$ & 66.46$_{\pm 0.80}$ & 56.93$_{\pm 1.21}$ 
    & 
    50.37$_{\pm 0.27}$ & 74.21$_{\pm 0.20}$ & 69.98$_{\pm 0.24}$\\
    \textproc{T5} & 
    53.49$_{\pm 0.13}$ & 74.22$_{\pm 0.08}$ & 72.36$_{\pm 0.12}$ &
    51.32$_{\pm 0.22}$ & 66.81$_{\pm 0.28}$ & 67.93$_{\pm 0.18}$ & 
    52.91$_{\pm 0.07}$ & 74.51$_{\pm 0.08}$ & 73.33$_{\pm 0.08}$\\
    
    \textproc{T5 + Fnn} &
    53.87$_{\pm 0.18}$ & 74.42$_{\pm 0.16}$ & 72.34$_{\pm 0.10}$& 
    50.71$_{\pm 0.12}$ & 66.42$_{\pm 0.24}$ & 67.06$_{\pm 0.24}$ & 
    52.71$_{\pm 0.14}$ & 74.32$_{\pm 0.11}$ & 73.32$_{\pm 0.16}$\\
    
    \textproc{T5-Graph} & 
    51.82$_{\pm 0.13}$ & 73.28$_{\pm 0.05}$ & 71.33$_{\pm 0.03}$&
    47.91$_{\pm 0.28}$ & 64.75$_{\pm 0.20}$ & 65.51$_{\pm 0.31}$ & 
    51.40$_{\pm 0.22}$ & 73.78$_{\pm 0.13}$ & 72.15$_{\pm 0.08}$\\
    \textproc{T5-Graph + Fnn} &
    52.04$_{\pm 0.22}$ & 73.58$_{\pm 0.15}$ & 71.37$_{\pm 0.13}$ &
    47.45$_{\pm 0.33}$ & 64.60$_{\pm 0.25}$ & 65.69$_{\pm 0.31}$ & 
    51.35$_{\pm 0.21}$ & 78.78$_{\pm 0.14}$ & 72.32$_{\pm 0.12}$\\
    \midrule
    \textproc{Ugt} & 
    54.27$_{\pm 0.24}$ & 75.13$_{\pm 0.10}$ & 72.13$_{\pm 0.16}$ &
    52.48$_{\pm 0.43}$ & 67.96$_{\pm 0.45}$ & 67.19$_{\pm 0.72}$ & 
    53.03$_{\pm 0.37}$ & 75.38$_{\pm 0.11}$ & 73.18$_{\pm 0.13}$\\
    \rowcolor[RGB]{237,237,237} \textproc{Ugt + Nse} & 
    \bf 55.95$_{\pm 0.23}$ & \bf 76.10$_{\pm 0.06}$ & \bf 72.84$_{\pm 0.18}$ &
    \bf 57.10$_{\pm 0.42}$ & \bf 71.74$_{\pm 0.43}$ & \bf 68.40$_{\pm 0.23}$ & 
    \bf 54.21$_{\pm 0.17}$ & \bf 75.93$_{\pm 0.20}$ & \bf 74.04$_{\pm 0.08}$\\
    \bottomrule
    \end{tabular}
    }
    \caption{Automatic evaluation results on the development and test sets. Mean ($\pm$s.d.) over 4 seeds.}
    \label{tab:main_results}
\end{table*}
\paragraph{Baselines}
Due to current datasets bias in the English language, the D2T methods for others are rarely explored. 
Meanwhile, PLMs-based models, such as T5, have achieved SOTA results~\citep{DBLP:journals/corr/abs-2007-08426,ribeiro-etal-2021-structadapt, DBLP:conf/naacl/WangXSC22,DBLP:conf/aaai/JollyZ0M22} on many D2T tasks. Therefore, we experiment with T5-based models to understand their performance on \catsd, \catss, and \cats:
\begin{itemize}
    \setlength{\itemsep}{0pt}
    \setlength{\parsep}{0pt}
    \setlength{\parskip}{0pt}
     \item \textproc{TemP} automatically generates descriptions based on the predefined template. Specifically, we first manually write a template for SQL queries replacing the values, columns, table names, and conditions with slots. Meanwhile, we also create a list of descriptions for each component in SQL queries (Table~\ref{sql_description} reports the descriptions of partial SQL components). Then we enumerate all cells in the table row by row to obtain the description for a table. Lastly, we join the two parts of descriptions as the final output.
     
    \item\textproc{Pointer-Gen} is an RNN-based Seq2Seq model with attention and copy mechanism \citep{DBLP:journals/corr/SeeLM17}. We concatenate the SQL and linearized table as input.
    \item \textproc{T5} denotes finetuning the T5 model on the proposed \cats. The input is the same as that used in the \textproc{Pointer-Gen}.
    Notably, to make a fair comparison with our proposed method, we add a fully connected feed-forward network (\textproc{Fnn}) on top of each transformer layer and make its parameters equal with the L-NE layer. We denote this as \textproc{T5 + Fnn}.
    \item \textproc{T5-Graph} is also a finetuning T5 method. Different from \textproc{T5}, it uses the sample graph representation with our method (described in Section \ref{subsec: unified}) as input.
    Again, we add FNN to make a fair comparison, which is denoted as \textproc{T5-Graph + Fnn}. 
\end{itemize}

\paragraph{Evaluation Metrics}We evaluate our models by applying both automatic and human evaluations. For automatic evaluation, we employ the widely used metric, BLEU \citep{papineni2002bleu} and ROUGE-L~\citep{lin2004rouge}, to evaluate the fluency of generated text. And we utilize SacreBLEU \citep{post2018call} to calculate the BLEU after segmenting the sentcne by jieba \footnote{http://pypi.python.org/pypi/jieba}. Additionally, we utilize \coverage~\citep{DBLP:conf/emnlp/ShaoHWXZ19} to evaluate the faithfulness of generated text. \coverage measures the average proportion of input tables that are covered by a generated text. The table headers are also considered. We use string matching rules to determine whether a cell exists in the generated text.
We conduct experiments over 4 different seeds and report the average scores on them.

We display examples of input representation for different models and provide the implementation details in Appendix~\ref{Appendixsec:example_for_model_inp} and ~\ref{Apendixsec:implement_details}.

\subsection{Main Result}
Table~\ref{tab:main_results} presents the experimental results on \cats, \catsd, and \catss, from which we make three main observations.

First, we can see that all neural network models outperform \textproc{TemP} on BLEU by a large margin. This suggests that neural models are better at generating fluent expressions. We consider this thanks to the language modeling task (Equation~\ref{equation_language_modeling}), which trains the neural models to predict the next token, given the previous history. Nevertheless, we find that \textproc{TemP} achieves the best \coverage scores on all sets, even better than \textproc{Gold}. We consider this is because, when annotating the references,  to make the presentation more reasonable and fluent, annotators summarize the contents of the table, such as merging some cells, etc. On the other hand, \textproc{TemP} copies all the contents of the table directly.

Second, adding extra trainable parameters (\textproc{+ Fnn}) does not always improve the performance on \textproc{T5} and \textproc{T5-Graph}. For example, \textproc{T5 + Fnn} performs better than \textproc{T5} on both \cats and \catss, but worse than \textproc{T5} on \catsd. 
Moreover, we notice that \textproc{T5} performs better than \textproc{T5-Graph} given the fact that the sizes of their parameters are equal.
We speculate this is because, compared to \textproc{T5-Graph}, \textproc{T5} uses the original SQL and the flattened table as input, which preserves the partial structural information of the input SQL and table by the segment symbols ``$,$'' and ``\textit{|}'' (please refer to Appendix~\ref{Appendixsec:example_for_model_inp} for the example of input data linearizations).
However, \textproc{T5-Graph} still treats the input as a sequence and ignores the unified graph's structure, leading to its performance degradation.

Lastly, by explicitly modeling the unified graph structures, \textproc{Ugt} dramatically outperforms the size-comparable models \textproc{T5-Graph + Fnn}  and \textproc{T5-Fnn} on all metrics.
The results display \textproc{Ugt}'s superiority in capturing essential structural knowledge for this task. 
Additionally, Node Segment Embedding (+ NSE) further improves the performance.
This verifies that \textproc{Nse} can help the encoder better preserve the original structural information. 
\subsection{Analysis and Discussion}

\paragraph{Effects of input SQL and Table}
To examine the effects of different input data, we conduct ablation studies on the input side by removing the input SQL and table. 
The results on three development sets are summarized in Table~\ref{tab:effect_of_input_data}. We observe that, after removing the SQL and only utilizing the table as input, both \textproc{T5 + Fnn} and our method (\textproc{Ugt + Nse}) perform poorly on all metrics.  
The performance degrades even more if only SQL is employed. 
The results demonstrate that both input SQL and table are essential for the answer-to-sequence task. 
Additionally, our method clearly outperforms \textproc{T5 + Fnn} on all ablation settings.
It reveals the effectiveness of our method compared to vanilla T5 architecture even under extreme input conditions. 

\begin{table}[t!]
    \centering
\resizebox{\columnwidth}{!}{
\begin{tabular}{lccc}
\toprule
\bf \textsc{Model} & \bf \cats & \bf \catsd & \bf \catss \\
\midrule
\rowcolor[RGB]{237,237,237} \textproc{T5 + Fnn} & 54.14$_{\pm 0.21}$ & 52.10$_{\pm 0.17}$ & 51.67$_{\pm 0.22}$ \\
\quad w/o \textproc{Sql} & 40.90$_{\pm 0.24}$ & 39.75$_{\pm 0.08}$ & 40.00$_{\pm 0.30}$ \\
\quad w/o \textproc{Table} & 17.83$_{\pm 0.13}$& 24.25$_{\pm 0.33}$ & 14.51$_{\pm 0.11}$ \\
\midrule
\rowcolor[RGB]{237,237,237} \textproc{Ours} & 56.34$_{\pm 0.13}$ & 58.79$_{\pm 0.51}$ & 53.54$_{\pm 0.15}$ \\
\quad w/o \textproc{Sql} & 45.16$_{\pm 0.26}$ & 47.92$_{\pm 0.50}$ & 43.89$_{\pm 0.38}$ \\
\quad w/o \textproc{Table} & 19.59$_{\pm 0.16}$ & 26.91$_{\pm 0.11}$ & 16.20$_{\pm 0.62}$ \\
\bottomrule
\end{tabular}
}
    \caption{Effect of input \textproc{Sql} and \textproc{Table}. w/o \textproc{Sql} and w/o \textproc{Table} denote removing the SQL and table from the input, respectively. \textproc{Ours} denotes~\textproc{Ugt + Nse}. }
    \label{tab:effect_of_input_data}
\end{table}
\paragraph{Effects of Data Complexity}
We further explore the performances on different levels of data complexity. 
We use BLEU as the metric in this section. 
The results are shown in Table \ref{tab:data_properties}. 
We first explore the effect of the table size. 
Unsurprisingly, the BLEU scores of all models decrease as the number of table rows or columns grows. 
The more rows or columns the table contains, the more difficult it is for a model to process it. 
Compared to two baseline models, our method is better at handling large tables. 
Furthermore, we investigate the impact of SQL complexity on model performances. 
With respect to the SQL complexity, our model achieves larger improvement against baseline models, especially on extra hard SQLs.
It shows that our approach can better encode the complex input data than others. 
Lastly, we study the model performance concerning different ground-truth description lengths. 
The \textproc{Pointer-Gen} struggles on longer descriptions, where the performance drops over 10 BLEU scores on responses longer than 60.
In this scenario, T5-based models dramatically outperform the \textproc{Pointer-Gen}, while our method can still beat \textproc{T5 + Fnn}.

\begin{table}[t!]
    \centering
\resizebox{\columnwidth}{!}{
\begin{tabular}{lcccc}
\toprule
\bf \textsc{Column Number} & 1& 2 & 3 & $>=$4 \\
\bf \textsc{\# Examples} & 1,138 & 2,580 & 403 & 215 \\
\midrule
\textproc{Pointer-Gen} & 53.21 & 50.74 & 42.20 & 35.29 \\
\textproc{T5 + Fnn}     & +2.28 & +1.16 & +7.08 & +4.29 \\
\rowcolor[RGB]{237,237,237} \textproc{Ours}             & \bf +5.61 & \bf +4.69 & \bf +7.54 & \bf +5.28 \\
\midrule
\bf \textsc{Row Number} & 1 & 2 & 3 & $>=$4 \\
\bf \textsc{\# Examples} & 1,899 & 769 & 467 & 1201  \\
\midrule
\textproc{Pointer-Gen} & 56.72 & 49.71 & 49.05 & 44.30 \\   
\textproc{T5 + Fnn}      & +3.57 & -0.58 & +1.68 & +6.24 \\
\rowcolor[RGB]{237,237,237} \textproc{Ours}              & \bf +5.75 & \bf +1.54 & \bf +5.16 & \bf +7.62  \\
\midrule
 \bf \textsc{SQL Hardness} & Easy & Medium & Hard & Extra Hard \\
 \bf \textsc{\# Examples} & 915 & 1,588 & 1,531 & 302 \\
\midrule
\textproc{Pointer-Gen} & 60.92 & 54.99 & 42.78 & 43.17 \\
\textproc{T5 + Fnn}      & +0.92 & +0.60 & +6.79 & +3.65 \\
\rowcolor[RGB]{237,237,237} \textproc{Ours}              & \bf +3.98 & \bf +3.75 & \bf +7.80 & \bf +9.22 \\
\midrule

 \bf \textsc{Target Length} & $<20$ & $<40$ & $<60$ & $>=60$ \\
\bf \textsc{\# Examples} & 1,275 & 1,635 & 724 & 702 \\

\midrule
\textproc{Pointer-Gen} & 52.67 & 51.97 & 52.02 & 41.64 \\
\textproc{T5 + Fnn}     & +2.93 & -0.31 & -0.06 & +7.54 \\
\rowcolor[RGB]{237,237,237} \textproc{Ours}         & \bf +6.08 & \bf +3.19 & \bf +3.33 & \bf +7.82 \\
\bottomrule
\end{tabular}
}
    \caption{BLEU scores of different models in the \cats test set on different levels of data complexity. Relative results of \textproc{T5 + Fnn} and our method are reported compared against the \textproc{Pointer-Gen}.}
    \label{tab:data_properties}
\end{table}

\subsection{Human Evaluation}
\label{sec:human_evaluation}
To reach a deeper understanding of the qualities of the generated descriptions, we conduct human evaluation following \citet{parikh2020totto}. We compare our method with \textproc{TemP}, \textproc{Pointer-Gen}, and \textproc{T5 + Fnn}. Specifically, we first randomly select 100 examples from the \cats test set and the corresponding outputs generated by each model. 
And then, five native Chinese annotators (three females and two males) with master's degrees or above engaged in NLP research are invited to evaluate the quality from the four axes. Specifically, \textbf{\textsc{Fluency}} measures whether the description is fluent. \textbf{\textsc{Faithfulness}} estimates whether the description is logically consistent with input SQL, and all pieces of information are supported by the input table. They are scores range from 1 to 10, the higher the better.
\textbf{\coverage} is the percentage of cells in the input table the candidate sentence covers. It is different from the one in Table~\ref{tab:main_results} (please refer to Appendix~\ref{Appendixsec:diff_coverage}).
\textbf{\textsc{Repetition}} is number of cells the candidate sentence repeats.
We also introduce the reference as one candidate (denoted as \textproc{Gold}). And its results can be regarded as the upper bound.

\begin{table}[t!]
    \centering

\resizebox{\columnwidth}{!}{
\begin{tabular}{lcccc}
\toprule
\bf \textsc{Model} & \bf Flu. $\uparrow$ & \bf Fai. $\uparrow$ & \bf Cov.(\%)$\uparrow$ & \bf Rep. $\downarrow$ \\
\midrule
\textproc{Gold} & 8.42 & 9.15 & 95.32 & 0.14 \\
\textproc{TemP} & 5.27 & 6.87 & 99.41 & 0.02 \\
\midrule
\textproc{Pointer-Gen} & 6.13 & 6.32 & 83.27 & 0.74 \\
\textproc{T5 + Fnn} & 6.82 & 7.16 & 89.27 & 0.39 \\
\rowcolor[RGB]{237,237,237} \textproc{Ours} & \bf 7.14 & \bf 7.48 & \bf 90.26 & 0.27 \\
\bottomrule
\end{tabular}
}
    \caption{Human evaluation over references (denoted as \textproc{Gold}) and model outputs. Flu., Fai., Cov., Rep. denote \textsc{Fluency}, \textsc{Faithfulness}, \textsc{Coverage} and \textsc{Repetition}. $\uparrow$ indicates higher is better and $\downarrow$ denotes lower is better.}
    \label{tab:human_evaluation}
\end{table}

The results summarized in Table~\ref{tab:human_evaluation} show that the \textproc{Gold} consistently achieves high performance than generation methods. It attests to the high quality of our human annotations. We report \textsc{Fluency} and \textsc{Faithfulness} score for \textproc{TemP} because they are sensitive evaluation. We can see that \textproc{TemP} gets a high \textsc{Faithfulness} score but is poor on \textsc{Fluency}. Our method outperforms baseline models on almost all axes with an agreement kappa score \citep{van2020human} more than $0.86$. It demonstrates the effectiveness of our proposed method. Although our model achieves a high coverage rate (90.26\%), its \textsc{Faithfulness} score is relatively low (only 7.48), and there is a considerable gap compared with \textproc{Gold}. It indicates simply copying content from the input table can not guarantee the faithfulness of the generated response. It may be necessary for the model to understand the deep semantics of SQL and table, which is the biggest challenge in this dataset.

\section{Conclusion}
We present \cats, a large-scale and high-quality Chinese answer-to-sequence dataset, along with a series of baselines. It helps alleviate the problem of current D2T datasets' bias towards the English language. 
We propose a Unified Graph Transformation method to bridge the structural gap between the SQL and table. 
In this way, we convert the task to a graph-to-text problem. 
Furthermore, we introduce the Node Segment Embedding to solve the problem that transforming the input graph to a new token graph breaks the original graph's structure.  
Experiments on \cats show that our proposed model outperforms existing baseline models. 
We conduct further analysis on \cats, which attests to both the high quality and challenges of the dataset.

\section*{Limitations}

This work presents \cats, a large-scale and high-quality Chinese answer-to-sequence dataset. It is a free and open dataset. One of most important motivations for presenting this dataset is that most of the existing datasets are built for English, which leads to advanced work on D2T generation primarily focusing on English and leaving other languages underexplored. However, \cats only alleviates the dataset language bias rather than solving it. And it is limited to the study of Chinese methods. Regarding methodology, the proposed \textproc{Ugt} converts the answer-to-sequence task to a graph-to-text problem to bridge the gap between two heterogeneous input data (SQL and table). However, \textproc{Ugt} works only for answer-to-sequence task rather than graph-to-text task. Additionally, though the proposed \textproc{Nse} can help the graph-to-text model better preserve the original structural information, the contribution may be limited to the graph-to-text task.

\section*{Ethics Statement}
This work presents \cats, a free and open dataset for the research community to study the answer-to-sequence problem in the practical TableQA system. And it helps enrich the D2T languages and alleviate the datasets' bias in English. 
To balance the data quality and scale and bring it closer to the practical scenario, data in \cats are collected from two sources, which are manually annotated as \catsd and \catss. 
In other words, \cats consists of \catsd and \catss.
The data in \catsd is collected from DuSQL~\citep{wang-etal-2020-dusql} dataset, a free and open dataset for the Chinese Text-to-SQL problem. Meanwhile, to enlarge our dataset, we adopt an automatic data construction pipeline to collect a large number of high-quality SQL-table pairs for annotation. To ensure the quality of our dataset, we manually annotate the SQL-table pairs. 
We hire 24 native annotators with undergraduate degrees to annotate the data.
Specifically, 20 annotators are responsible for annotations, and another 4 workers are asked to review the annotated data.
We pay 2.1 yuan (\$0.31 USD) for annotating each SQL-table pair.

To avoid our dataset leakages personal privacy, we replace the sensitive information in the collected tables with predefined unique words. Furthermore, we ask the annotators to filter out the examples that leak personal privacy and contain social bias and harmful content.

\bibliography{cats}
\bibliographystyle{acl_natbib}

\appendix
\clearpage
\begin{table}[t]
\centering
    \centering
    \resizebox{\columnwidth}{!}{
    \begin{tabular}{c}
        \toprule
        \texttt{SQLs ::= SQL | SQL interaction SQLs  |} \\ \texttt{SQL union SQLs | ... }\\
        \midrule
        \texttt{SQL ::= Select | Select Where | }\\ \texttt{ Select Order | Select Order Filter } \\
        \midrule
        \texttt{Select ::= Select A | Select AA | ... }\\
        \midrule
        \texttt{Where ::= Where Conditions} \\
        \midrule
        \texttt{Conditions ::= A op value | A op SQL} \\
        \midrule
        \texttt{A ::= C | MIN C | MAX C | AVG C | }\\ \texttt{COUNT C | SUM C} \\
        \midrule
        \texttt{C ::= T.column |  T.column mathop T.column} \\
        \midrule
        \texttt{T ::= table in current database} \\
        \midrule
        \texttt{mathop ::= + | - | * | \/} \\
        \midrule
        \texttt{op ::= == | != | > | >= | < | <= | like | in | not in}  \\
        \bottomrule 
    \end{tabular}
    }
    \caption{SQL generation grammar rules.}
    \label{sql_grammar}
\end{table}
\section{Dataset Construction Details}
\subsection{Database Building Details}
\label{table_clean} 
To build the database, we first clean the collected tables.
We build a rule-based table cleaning pipeline to guarantee table quality. We filter out noise tables via rules as follows: 
(1) We first build a blacklist including special chars, dirty words, emojis, and HTML words. And filter tables if the headers or the values include any word in the blacklist; 
(2) We recognize all of the header types in each table including \texttt{Text, Number, Time, and Bool}. If the proportion of \texttt{Text} type is less than 30\%, we filter out the table; 
(3) We filter out tables with less than 2 columns or rows; 
(4) We will filter out the table, if a value repeats more than 50\% in it. Finally, we obtain 24K high-quality tables.

The original crawled data are in the form of independent tables, which need to be linked with other tables to form databases. We build a database creation pipeline and link different tables based on the header overlap~\citep{wang-etal-2020-dusql} 
to acquire multi-table databases. Finally, 600 databases are selected in the dataset.

\subsection{Automatic SQL Generator}
\label{appendixsec:sql_generator}
The SQL generator utilizes production rules from the SQL grammar to automatically generate SQL queries. Specifically, a SQL query can be represented as an abstract syntax tree (AST) using the rules, such as \texttt{{SQLs = SQL, SQL = Select Where, Select = SELECT A, Where = WHERE Conditions...}}, all of which are production rules of the SQL grammar.
By exploiting every rule of the grammar, we can generate SQL queries covering patterns of different complexity along with corresponding tables. We illustrate some SQL production rules in Table~\ref{sql_grammar}.
\begin{figure}[t!]
\centering
\includegraphics[width=1.0\columnwidth]{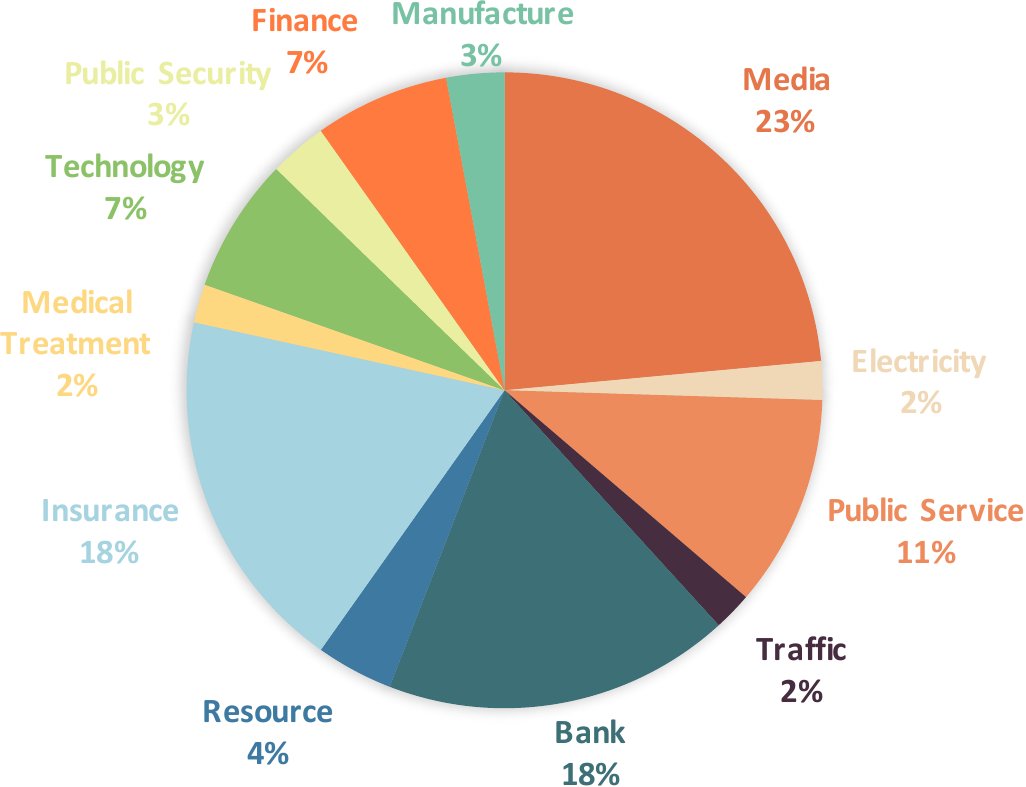}
\caption{Topic distribution of \cats.}
\label{fig:topic}
\end{figure}

\subsection{SQL Hardness}
Following~\citet{guo2021chase}, we adopt SQL hardness to measure the complexity of SQL queries from the following four-level: 
easy, medium, hard, and extra hard \citep{yu2018spider}. The SQL difficulty is defined based on the number of SQL components, selections, and conditions. Therefore, queries that contain more SQL keywords (GROUP BY, ORDER BY, INTERSECT, nested subqueries, column selections, and aggregators, etc.) are considered harder. For example, a query is considered hard if it includes more than two SELECT columns, more than two WHERE conditions, and GROUP BY two columns, or contains EXCEPT or nested queries. A SQL with more additions on top of that is considered extra hard.

\section{Topics Distribution of \cats}
\label{Appendixsec:cats_topic}
Following \citet{parikh2020totto}, we build a topic categorization model 
for tables in \cats to investigate the topic distribution. We first ask the annotators to label 10,000 tables and then train a table topic classifier built on a table-aware encoder~\cite{bao2018table}. We apply the classifier to label other table topics.
Figure~\ref{fig:topic} presents an aggregated topic analysis of our dataset. 
We find that 61\% of \cats is made up of the \texttt{Media}, \texttt{Insurance}, and \texttt{Bank topics}, and the other 39\% is composed of broader topics, such as \texttt{Public Service}, \texttt{Technology}, and \texttt{Finance}.
The proposed \cats is limited to topics that are presented in CLUE and DuSQL.

\section{Experimental Details}

\subsection{Example of SQL and Table Linearizations}
\label{Appendixsec:example_for_model_inp}
\begin{figure}[t]
\centering
\includegraphics[width=1.0\columnwidth]{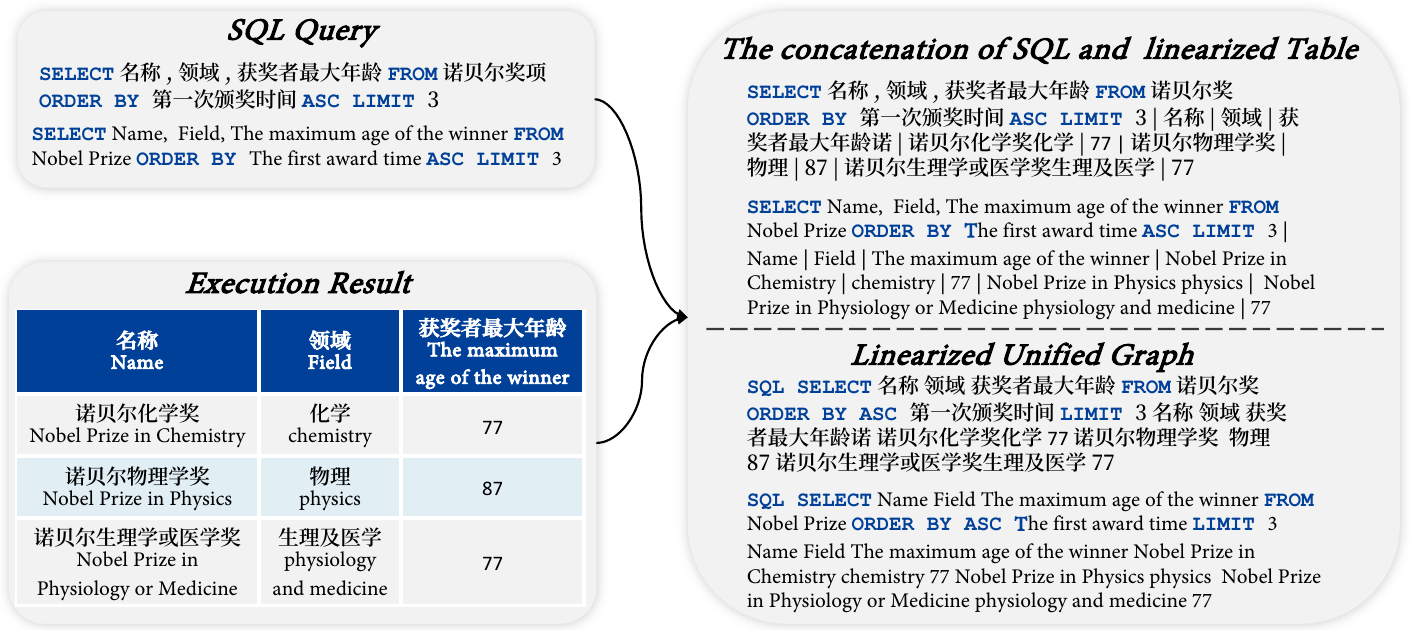}
\caption{Different linearizations for an input SQL and Table.}
\label{fig1:inp_rep_for_diff_models}
\end{figure}

We display the input representations for different models in Figure~\ref{fig1:inp_rep_for_diff_models}.
For \textproc{Pointer-Gen}, \textproc{T5}, and \textproc{T5 + Fnn}, we directly concatenate the SQL and linearized table as input, where table is linearized row by row. For \textproc{T5-Graph}, \textproc{T5-Graph + Fnn} and \textproc{Ours}, follow previous work~\citep{ribeiro-etal-2021-structadapt}, we linearize the SQL graph $\mathcal G_s$ into a sequence of nodes by the depth-first traversal and concatenate it with the linearized table as input. Especially, instead of segmenting the nodes with special symbol \textit{|}, we build a connection matrix for the token graph $\mathcal G$. The connection matrix is used by the Local Node Encoder to encoding the graph structure.

\subsection{Implementation Details}
\label{Apendixsec:implement_details}
We employ the \textproc{Pointer-Gen} implemented by OpenNMT \citep{klein-etal-2017-opennmt}. \textproc{Pointer-Gen} is built based on LSTM~\citep{DBLP:journals/neco/HochreiterS97}. We set the layers of the encoder and decoder as $2$ and $1$, respectively. And we set the embedding and decoder hidden size as $512$. T5-based methods are implemented using HuggingFace \citep{wolf-etal-2020-transformers} and inintilized by T5$_{base}$\footnote{https://huggingface.co/uer/t5-base-chinese-cluecorpussmall}. And the hidden size of the GAT layer in the Local Node Encoder is set to $512$. For T5-based methods, we set the dropout rate to $0.1$, use AdamW optimizer \citep{loshchilov2018decoupled} and employ a linear learning rate decay schedule without warm-up. We use BLEU \cite{papineni2002bleu} for the early stopping criterion. Moreover, the learning rate is 3e-5 and batch size is 4 for all experiments. During decoding, we employ beam search with a beam size $5$. All experiments are trained on Nvidia Tesla V100 32GB GPUs.

\subsection{Human Evaluation Details}
The detailed information about the four human evaluation metrics are as following:
\begin{itemize}
    \item \textbf{Fluency}: a sentence is fluent if it is grammatical and natural. And it is scored from 1 to 10, where 1 represents not Fluent, and 10 represents Mostly Fluent.

    \item \textbf{Faithfulness}: a sentence is considered faithful if it is logically consistent with the input SQL and all pieces of information are supported by the table. The score ranges from 1 to 10.

    \item \textbf{Coverage} is the percentage of cells in the input table the candidate sentence covers. It is calculated by $\frac{n^c}{n^t}$, where $n^t$ denotes all cells in the input table, and $n^c$ represents the number of cells covered by the sentence.

    \item \textbf{Repetition} number of cells the candidate sentence repeats. If a cell is repeated $n$ times, it will be recorded $n$ times.
\end{itemize} 

For each sample, the annotators need to evaluate four candidates based on the input data. And they do not know which model generates these sentences.

\subsection{Differences in \coverage between Automatic Evaluation and Human Evaluation}
\label{Appendixsec:diff_coverage}
The \coverage in Table~\ref{tab:main_results} is calculated by $cov^{a} = \frac{n^c}{n^a}$, where $n^a$ denotes all cells in the input table and include the cells in the table header. $n^c$ represents the number of cells covered by the generated text. We use string matching rules to determine whether a cell exists in the generated text. $cov^{a}$ does not consider semantic matching between cells. Therefore, it will miss some cells that are summarized or paraphrased cells.

The \coverage in human evaluation is calculated $cov^{h} = \frac{n^c}{n^t}$, where $n^t$ denotes all cells in the input table and does not include the cells in the table header. $n^c$ represents the number of cells covered by the sentence. $n^c$ is counted by manual checking. Therefore, the cells that are summarized or paraphrased in the generated text will counted.

Overall, $cov^{a}$ is more rigorous and inflexible than $cov^{h}$, and it takes more account of the able headers, so it scores lower.

\section{Case Study}
\begin{figure}[!ht]
    \includegraphics[width=1.0\columnwidth]{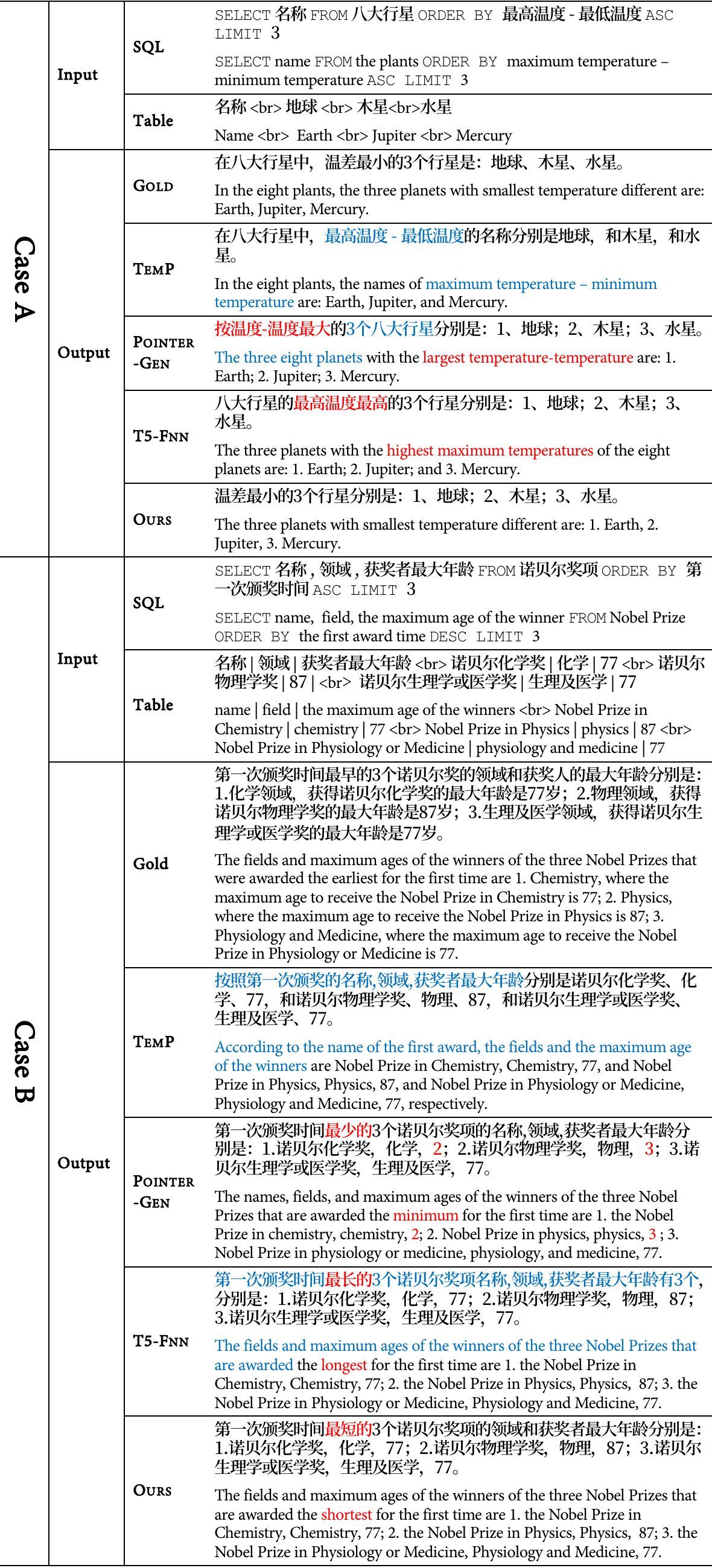}
    \caption{Answer-to-sequence examples in \cats. Error words are in red. Confusing and incomprehensible phrases are in blue.}
    \label{fig:case_study}
\end{figure}

In Figure~\ref{fig:case_study}, we display two decoder output examples from the baselines on the development set of \cats. We find that the model can generate text with high coverage when the input table is simple, such as the number of columns being small. Second, when the input table is complex, such as containing multiple rows and columns, simple models, such as \textproc{Pointer-Gen}, tend to miss some content. Meanwhile, the complex models, such as T5-based ones, only simply enumerate the table cells rather than describe them like humans. Finally, the descriptions generated by models are not faithful to the input, even though they contain most of the input table content. For example, in the second case, all the models do not describe the ``earliest'' correctly. That is, the descriptions are not logically consistent with the input SQL, which is one of the biggest challenges of this task.

\end{document}